\documentclass{IOS-Book-Article}
\usepackage[utf8]{vietnam} 
\usepackage{mathptmx}
\usepackage{soul}\setuldepth{article}
\usepackage[english]{babel}
\usepackage{amsmath}
\usepackage{multirow}
\usepackage{float}
\usepackage{threeparttable} 
\usepackage{amsmath}
\usepackage{enumerate}
\def\hb{\hbox to 10.7 cm{}}

\begin{document}

\pagestyle{headings}
\def\thepage{}

\begin{frontmatter}              

\title{Vietnamese Complaint Detection on E-Commerce Websites\\}

\markboth{}{April 2020\hb}

\author[A,B]{\fnms{Nhung Thi-Hong Nguyen}},
\author[A,B]{\fnms{Phuong Phan-Dieu Ha}},
\author[A,B]{\fnms{Luan Thanh Nguyen}},
\author[A,B]{\fnms{Kiet Van Nguyen}},
\author[A,B]{\fnms{Ngan Luu-Thuy Nguyen}}

\runningauthor{B.P. Manager et al.}
\address[A]{University of Information Technology, Ho Chi Minh City, Vietnam}
\address[B]{Vietnam National University Ho Chi Minh City, Vietnam}

\begin{abstract} \textbf{. }
Customer product reviews play a role in improving the quality of products and services for business organizations or their brands. Complaining is an attitude that expresses dissatisfaction with an event or a product not meeting customer expectations. In this paper, we build a \textbf{Vi}etnamese \textbf{O}pen-domain \textbf{C}omplaint \textbf{D}etection dataset (UIT-ViOCD), including 5,485 human-annotated reviews on four categories about product reviews on e-commerce sites. After the data collection phase, we proceed to the annotation task and achieve the inter-annotator agreement (${A_{m}}$) of 87\%. Then, we present an extensive methodology for the research purposes and achieve 92.16\% by F1-score for identifying complaints. With the results, in future, we aim to build a system for open-domain complaint detection on E-commerce websites.

\keywords{Customers' Complaint \and Vietnamese Dataset \and Deep Learning \and Transfer Learning}
\end{abstract}

\end{frontmatter}
\markboth{April 2020\hb}{April 2020\hb}

\section{Introduction}
In the era of technology development, the trend of online shopping on e-commerce sites is rapidly increasing. Consumers can easily express their opinions or feelings by leaving comments to resolve the complaint issues. Manually processing responses takes a significant amount of time and effort, hence identifying complaints should be automated. According to Olshtain and Weinbach
1987 \cite{olshtain198710}, complaining is a basic speech act used to express a negative mismatch between the expectation and reality towards a state of affairs, products, organizations, or events.  

Datasets used for complaint detection are now not so much, especially in Vietnam; there is no dataset about that so far. Therefore, in this paper, we present a human-annotated dataset for this task to automatically identify complaints and contribute to a more robust Vietnamese dataset. After building the dataset, we implement machine learning-based methods as baseline systems to evaluate our dataset.

Our contributions in this paper include:

\begin{itemize}
    \item We introduce a Vietnamese dataset named UIT-ViOCD, which contains 5,485 human-annotated reviews for identifying complaints on the open-domain such as fashion, cosmetics, applications, and phones.
    
    \item We conduct multiple experiments on different machine learning-based models, including traditional machine learning, deep learning, transfer learning models with low to high complexity and analyze the quantitative dataset. The highest performance is reached by PhoBERT with 92.16\% F1-score for complaint speech detection. We also implement different experiments to figure out the role of Vietnamese structure ingredients.
\end{itemize}

We organize the rest of the paper as follows. Section 2 presents the works related to our task. The process of building and analyze the dataset is described in Section 3. In Section 4, we propose our methodology and discuss experimental results. Finally, we draw a conclusion and future work in Section 5.

\section{Related Work} 
In 1987, Olshain et al. \cite{olshtain198710} stated that complaints are the result when a speaker is adversely affected or expects a favorable event to occur. Complaints can also be described in a positive sense as a report from a consumer documenting a problem with a product or service \cite{felix2016effectiveness}. Complaints usually co-occur with other speech acts such as warnings, threats, suggestions, or advice. Moreover, Cohen et al. \cite{cohen1993production} have found that the complaint often occurs as "speech act set".

In the world, there are various studies about customer complaints in different languages. In 2002, Pang et al. \cite{pang2002thumbs} used supervised machine learning methods to categorize film reviews with datasets available from the rec.arts.movies.reviews archive. In 2004, Hu and Liu \cite{hu2004mining} did the research with an aim to mine and summarize all the customer reviews of products, namely, they mined the features of the products collected from the reviews on Amazon.com and C|net.com, on which the customers have expressed their opinions positively or negatively. Moreover, Gaman and N. Aletras used Distant Supervision, deep learning methods to identify Twitter complaints in English with nine domains \cite{preotiuc2019automatically}.

In Vietnamese, the current datasets mostly focus on the emotional recognition problem. Minh et al. researched aspect-based emotional analysis and published a dataset including 7,828 restaurant reviews for solving two sub-tasks: aspect detection and polarity detection \cite{nguyen2019corpus}. In 2019, Vong et al. built a standard Vietnamese Social Media Emotion Corpus (UIT-VSMEC) consisting of 6,927 human-annotated comments with six emotion labels \cite{ho2019emotion}. In 2021, Nguyen et al. published a Vietnamese dataset about constructive and toxic speech detection named UIT-ViCTSD \cite{nguyen2021constructive}.  Nonetheless, there is no dataset for identifying complaints so far. Therefore, we decide to build a dataset about customer complaints about automatically classifying comments into complaints or non-complaints.
\section{Dataset}
We collect customer product reviews from e-commerce sites and Google Play Store\footnote{https://play.google.com/store} with 5,485 reviews belonging to four areas: fashion, cosmetics, applications, and phones. 
\subsection{Task Definition And Guidelines}
Based on the complaint definition proposed by Olshtain and Weinbach \cite{olshtain198710}, we build annotation guidelines for our task.

\textbf{Complaint (Label 1):} Complaining reviews express customers of dissatisfaction between reality and expectations, often accompanied by the build-up and desire to resolve. Complaints usually co-occur with other speech acts such as warnings, threats, suggestions, or advice \cite{preotiuc2019automatically}. Besides, complaining reviews sometimes contain compliments in addition to complaints. Complaints are constructive, not showing hateful emotions or heavy criticism regarding the service or product. Punctuation and special characters, icons can also be physical signs that show complaints.

\textbf{Non-complaint (Label 0):} Non-complaining reviews are compliments, praise comments, satisfaction about products and services. 

\begin{table}[hbt!]
\centering
\caption{Examples of reviews annotated for complaint task} 
\label{tab:examples_label}
\begin{tabular}{clcc}
\hline
\multicolumn{1}{l}{\textbf{No}} & \multicolumn{1}{c}{\textbf{Comment}}                                                                                                                           & \textbf{Label} \\ \hline
1                                & \begin{tabular}[c]{@{}l@{}}Phải chi có tiếng việt nữa thì tuyệt vời.\\ \textit{(It's great to have Vietnamese.)}\end{tabular}                                           & 1\\
2                                & \begin{tabular}[c]{@{}l@{}}Chất lượng rất tốt, thời gian xử lý đơn hàng và giao hàng quá lâu.\\ \textit{(Very good quality, but take too much time for processing and delivery.)}\end{tabular}                                           & 1\\
3                                & \begin{tabular}[c]{@{}l@{}}Không thấy quà tặng theo là như nào lừa đảo à ???\\ \textit{(Don't see gifts, scam ???)}\end{tabular}                                        & 1 \\
4                                & \begin{tabular}[c]{@{}l@{}}Game như L** đừng có tải nha\\ \textit{(Game likes P*ssy don't download it)}\end{tabular}                                                           & 0\\
5                                & \begin{tabular}[c]{@{}l@{}}Mình đã sd đến chai t5 của srm senka r rất  ưnggg\\ (\textit{I have used the 5th bottle of senka cleanser so satisfied)}\end{tabular} &                   0\\ \hline
\end{tabular}
\end{table}

Table \ref{tab:examples_label} presents examples of reviews highlighting differences between complaints and non-complaints in our dataset. In the first sentence, the assessment sentence, which sets out unmet customer wishes but does not show negative sentiment, is constructive. The second sentence has both praise and dissatisfaction and is classified as a complaint. Meanwhile, the third sentence shows mixed negative sentiment (angry emotion) and expressed a complaint. In the fourth sentence, we see an insult that implies negative sentiment but is not constructive for the business organizations; hence, this sentence can not be labeled as a complaint. The fifth sentence shows a positive and satisfied attitude to the product.

\subsection{Annotation Process}
It is necessary to know the task for building a high-quality and reliable dataset, thus we create a guideline for annotators before labeling data. The annotator team consists of 3 people who have enough knowledge about natural language processing and labeling data. For evaluating the inter-agreement among annotators, we use the Fleiss' Kappa \cite{bhowmick2008agreement}, a statistical technique according to the formula (1). 
\textit{}\begin{center}$$\mathrm{A}_{\mathrm{m}}=\frac{\mathrm{P}_{o}-\mathrm{P}_{e}}{1-\mathrm{P}_{e}} \quad\quad(1)$$  \end{center}

Where, the observed agreement ${P_o}$ is the proportion of sentences in which both of the annotators agreed on the class pairs. The chance agreement ${P_e}$ is the proportion of items for which agreement is expected by chance when the sentences are seen randomly. First, we label a part of the dataset. Then, we measure the ${A}_{m}$ and update guidelines. We repeat this process until the inter-agreement of annotators reaches over 80\%. After getting that score, annotators have enough knowledge about the task, and we begin labeling the rest of the data consecutively. 

\subsection{Dataset Analysis}
There is no significant difference between 2,854 complaining reviews (52.03\%) and 2,631 non-complaining reviews (47.97\%). 


To have a deeper understanding of the UIT-ViOCD dataset, we analyze the distribution of the complaints dataset according to the length of the sentence, and the results are shown in Table \ref{tab:len_of_sentence}. The number of sentences whose length is greater than 30 and makes up most of the dataset, especially in Label 1, occupies up to 46.45\%. It shows that customers aim to express their desires and opinions in detail and exact.

\begin{table}[hbt!]
\centering
\caption{Distribution of the complaint-annotated dataset according to the length of the sentence (\%).}
\label{tab:len_of_sentence}
\begin{tabular}{cccc}
\hline
\textbf{Length}           & \textbf{Label 0} & \textbf{Label 1} & \textbf{Overall} \\ \hline
\textbf{1-5}              & 0.40              & 0.04             & 0.44           \\
\textbf{6-10}             & 1.42             & 0.27             & 1.69           \\
\textbf{11-15}            & 4.41             & 0.86             & 5.27           \\
\textbf{16-20}            & 3.94             & 1.22             & 5.16           \\
\textbf{21-25}            & 2.94             & 1.22             & 4.16           \\
\textbf{26-30}            & 3.43             & 1.97             & 5.40            \\
\textbf{\textgreater{}30} & \textbf{31.43}            & \textbf{46.45 }           & \textbf{77.88}          \\ 
\textbf{Total}            & 47.97             & 52.03             & 100.00            \\
\hline
\end{tabular}
\end{table}

Most customers tend to complain about quality, delivery service, and price. The application field is slightly different from other areas, and customers often show a positive attitude to improve the application. The rest of the sectors tend to be straight to offer dissatisfaction.

We perform a vocabulary-based analysis experiment to understand linguistic features between complaints and non-complaints. This experiment is Part-of-speech, Word Frequency Dictionary, and Embedding Extraction from the identifying complaint task. We assign part-of-speech to each word in every review of our dataset, such as nouns, verbs, adjectives, using ViPosTagger (Vietnamese pos tagging F1\_score = 92.5\%) of Pyvi library\footnote{https://pypi.org/project/pyvi/}. We create a frequency dictionary to map the word and the class it appeared to in (complaint and non-complaint) to the number of times that word appeared in its corresponding class. Top part-of-speech features that belong to complaints and non-complaint are analyzed about word feature similarity using Cosine similarity \cite{li2013distance} (2). Where, $\vec{a}$ and $\vec{b}$ are two vectors of two words extracted from the embedding word model of the this task. Table  \ref{big_table} shows the results of this experiment.

$$\operatorname{Cos} (\vec{a},  \vec{b})=\frac{\vec{a} \cdot \vec{b}}{\|\vec{a}\|\|\vec{b}\|}=\frac{\sum_{1}^{n} a_{i} b_{i}}{\sqrt{\sum_{1}^{n} a_{i}^{2}} \sqrt{\sum_{1}^{n} b_{i}^{2}}}     \quad\quad(2)$$

With the results in Table \ref{big_table}, linguistic features from part-of-speech are distinctive of complaints. In detail, several part-of-speech patterns such as coordinating conjunction and adverb are important Vietnamese complaints classification features. Linguistics research identified that a significant proportion of complaints tended to juxtapose overall negative evaluation with some positive appraisal \cite{vasquez2011complaints}. Definitely, "nhưng" (but) is the word that appears the most in complaining reviews and goes with the good and the bad of the product (e.g. "giao hàng chậm nhưng sản phẩm ok, nguyên seal.",  means: "slow delivery but the product is ok, fully sealed."). Moreover, "không" (not, no) often accompanies words that describe the service or product issue. Furthermore, "rất" (very) comes with good characteristics.

Besides, we realize that keywords with complaint meanings (top words) usually go with negative words. Likewise, words that go with non-complaint keywords often have not negative sense.
\begin{table}[hbt!]
\caption{Vietnamese abbreviations in the UIT-ViOCD dataset.}
\centering
\begin{tabular}{|c|c|c|}
\hline
\textbf{Abbreviations} & \textbf{Normalization} & \textbf{English meaning} \\ \hline
"dx", "dc", "dk" & "được" &"ok" \\
\hline
"dt" & "điện thoại" & "phone" \\
\hline
"god" & "tốt" & "good" \\
\hline
"ship" & "giao hàng" & "ship" \\ \hline
\end{tabular}
\end{table}
\begin{table}[]

\caption{The experiment results: Part-of-speech (POS), Word Frequency Dictionary, and Embedding Extraction from Identifying Complaint Task.}
\centering
\label{big_table}
\begin{tabular}{|c|c|l|c|}
\hline
\multicolumn{4}{|c|}{\textbf{Complaint}}                                                                                                                                                                                                                            \\ \hline
\textbf{POS}                        & \textbf{Top words}                                                   & \multicolumn{1}{c|}{\textbf{Words have high similarity scores to top-word}}                                                                                                                     & \textbf{Similarity score}                                        \\ \hline
\multirow{3}{*}{\textbf{Verb}}      & \begin{tabular}[c]{@{}c@{}}cập nhật\\ (update)\end{tabular}         & \begin{tabular}[c]{@{}l@{}}hôi cực (smell), trễ (late), lựa chọn (choose), xấu (bad),\\càng (more)\end{tabular}                                  & \begin{tabular}[c]{@{}c@{}}[0.294,  0.384]\end{tabular}  \\ \cline{2-4} 
                                    & \begin{tabular}[c]{@{}c@{}}khắc phục\\ (fix)\end{tabular}           & \begin{tabular}[c]{@{}l@{}}tẩy trang (cleansing), không thể (can’t), trôi (drift), thanh\\toán (payment), lâu (slow)\end{tabular}                                         & \begin{tabular}[c]{@{}c@{}}[0.070,  0.104]\end{tabular}  \\ \cline{2-4} 
                                    & \begin{tabular}[c]{@{}c@{}}thất vọng\\    (disappoint)\end{tabular} & \begin{tabular}[c]{@{}l@{}}mỏng (thin), nhưng (but), trừ (minus), đề nghị (suggest),\\tẩy trang (cleansing)\end{tabular}              & \begin{tabular}[c]{@{}c@{}}[0.089,  0.119]\end{tabular}  \\ \hline
\multirow{2}{*}{\textbf{Adjective}} & \begin{tabular}[c]{@{}c@{}}tệ \\    (bad)\end{tabular}               & \begin{tabular}[c]{@{}l@{}}sơ sài (incomplete), mỏng (thin), cồn (alcohol), rách (torn),\\chậm (slow), nhưng (but)\end{tabular}                               & \begin{tabular}[c]{@{}c@{}}[0.029,  0.082]\end{tabular}  \\ \cline{2-4} 
                                    & \begin{tabular}[c]{@{}c@{}}sai \\    (wrong)\end{tabular}            & \begin{tabular}[c]{@{}l@{}}lộn (mismatch), lừa đảo (cheat), tệ (bad), mỏng (thin), sơ\\sài (imcomplete), hư hỏng (broken), kém, xấu (bad),\\không (not)\end{tabular}    & \begin{tabular}[c]{@{}c@{}}[0.077,  0.125]\end{tabular}  \\ \hline
\multirow{2}{*}{\textbf{Noun}}      & \begin{tabular}[c]{@{}c@{}}lỗi \\    (error)\end{tabular}            & \begin{tabular}[c]{@{}l@{}}chậm (slow), gửi (send), móp (dented), bắt (must), thanh\\toán (payment), nhu cầu (need), chán (bored), khắc phục\\(fix)\end{tabular}         & \begin{tabular}[c]{@{}c@{}}[0.105,  0.124]\end{tabular}  \\ \cline{2-4} 
                                    & \begin{tabular}[c]{@{}c@{}}trò chơi\\ (game)\end{tabular}           & \begin{tabular}[c]{@{}l@{}}đề nghị (suggest), đéo (fuck), thất vọng (disappoint), không\\(not)\end{tabular}                                                               & \begin{tabular}[c]{@{}c@{}}[0.315,  0.377]\end{tabular}  \\ \hline
\multirow{2}{*}{\textbf{Others}}  & \begin{tabular}[c]{@{}c@{}}không\\ (not, no)\end{tabular}            & \begin{tabular}[c]{@{}l@{}}chậm (slow), gửi (send), móp (dented), bắt (must), thanh\\toán (payment), nhu cầu (need), chán (bored), khắc phục\\(fix)\end{tabular}         & \begin{tabular}[c]{@{}c@{}}[0.107,  0.174]\end{tabular} \\ \cline{2-4} 
                                    & \begin{tabular}[c]{@{}c@{}}nhưng\\ (but)\end{tabular}                & \begin{tabular}[c]{@{}l@{}}sơ sài (incomplete), rách (torn), gửi (send), mỏng (thin),\\thất vọng (disappoint), cồn (alcohol), chậm (slow), tệ\\(bad)\end{tabular}         & \begin{tabular}[c]{@{}c@{}}[0.082,  0.115]\end{tabular} \\ \hline
\multicolumn{4}{|c|}{\textbf{Non-Complaint}}                                                                                                                                                                                                                                                                                                               \\ \hline
\textbf{POS}                        & \textbf{Top words}                                                   & \multicolumn{1}{c|}{\textbf{Words have high similarity scores to top-word}}                                                                                                                     & \textbf{Similarity score}                                               \\ \hline
\multirow{2}{*}{\textbf{Verb}}      & \begin{tabular}[c]{@{}c@{}}thích \\    (like)\end{tabular}           & \begin{tabular}[c]{@{}l@{}}êm (smooth), kỹ lưỡng (elaborate), cảm ơn (thank), tiện\\dụng (comfortable)\end{tabular}                                & \begin{tabular}[c]{@{}c@{}}[0.294,  0.384]\end{tabular}  \\ \cline{2-4} 
                                    & \begin{tabular}[c]{@{}c@{}}cảm ơn\\ (thank)\end{tabular}            & \begin{tabular}[c]{@{}l@{}}xinh , dễ thương (lovely), nhờ (thanks to), yêu (love), rẻ\\(cheap), nhanh chóng (fast)\end{tabular}                & \begin{tabular}[c]{@{}c@{}}[0.071,  0.119]\end{tabular}  \\ \hline
\multirow{3}{*}{\textbf{Adjective}} & \begin{tabular}[c]{@{}c@{}}đẹp\\ (beautiful)\end{tabular}            & \begin{tabular}[c]{@{}l@{}}rẻ (cheap), biết (know), nhờ (thanks to), mua (buy), tiện\\(convenient), tốt, xịn (good)\end{tabular}                                            & \begin{tabular}[c]{@{}c@{}}[0.102,  0.137]\end{tabular}  \\ \cline{2-4} 
                                    & \begin{tabular}[c]{@{}c@{}}nhanh \\    (fast)\end{tabular}           & \begin{tabular}[c]{@{}l@{}}hài lòng (sastified), dễ thương (lovely), thích (like), thương\\(love), xuất sắc (excellent)\end{tabular}                                     & \begin{tabular}[c]{@{}c@{}}[0.079,  0.117]\end{tabular}  \\ \cline{2-4} 
                                    & \begin{tabular}[c]{@{}c@{}}đúng \\    (right)\end{tabular}           & \begin{tabular}[c]{@{}l@{}}vợ (wife), ưng ý (like), kỹ càng (elaborate), dễ thương\\(lovely), cảm ơn (thank)\end{tabular}                           & \begin{tabular}[c]{@{}c@{}}[0.189, 0.245]\end{tabular}  \\ \hline
\textbf{Noun}                       & \begin{tabular}[c]{@{}c@{}}sao \\    (star)\end{tabular}             & vui vẻ (happy), kĩ càng (elaborate)                                                                                                                          & \begin{tabular}[c]{@{}c@{}}[0.156,  0.166]\end{tabular}  \\ \hline
\textbf{Others}                   & \begin{tabular}[c]{@{}c@{}}rất \\    (very)\end{tabular}             & \begin{tabular}[c]{@{}l@{}}mình (I), êm (smooth), đi (working), điện thoại\\(smartphone), tuyệt vời (excellent), rẻ (cheap), xài (use),\\đầy đủ (enough)\end{tabular} & \begin{tabular}[c]{@{}c@{}}[0.270, 0.312]\end{tabular}  \\ \hline
\end{tabular}
\end{table}

\subsection{Comparison with Other Vietnamese Datasets}
Table \ref{tab:datasetcomp} presents many Vietnamese datasets for different tasks, which is enough to build information recognition features on the social listening system. Compared to other Vietnamese datasets, our dataset (UIT-ViOCD) is the first Vietnamese dataset for complaint classification.

\begin{table}[H]
\centering
\caption{Vietnamese datasets for evaluating different text classification tasks.}
\label{tab:datasetcomp}
\begin{tabular}{|l|r|l|r|l|}
\hline
\multicolumn{1}{|c|}{\textbf{Name}} & \multicolumn{1}{c|}{\textbf{Year}} & \multicolumn{1}{c|}{\textbf{Task}}             & \multicolumn{1}{c|}{\textbf{Size}} & \multicolumn{1}{c|}{\textbf{Domain}} \\ \hline
SA-VLSP2016 \cite{nguyen2018vlsp}                         & 2016                               & Sentiment Analysis                             & 12,196                             & Open-domain                          \\ \hline
SA-VLSP2018 \cite{nguyen2018vlsp}                         & 2018                               & Aspect-based Sentiment Analysis                & 10.351                             & Hotel and Restaurant                 \\ \hline
UIT-VSFC \cite{van2018uit}                            & 2018                               & Sentiment Analysis                             & 16,175                             & Education                            \\ \hline
UIT-VSMEC \cite{ho2019emotion}                           & 2019                               & Emotion Recognition                            & 6,927                              & Open-domain                          \\ \hline
ReINTEL2020 \cite{le2020reintel}                         & 2020                               & Responsible Information Identification         & 10,007                             & Open-domain                          \\ \hline
UIT-ViCTSD \cite{nguyen2021constructive}                          & 2021                               & Constructive and Toxic Detection & 10,000                             & Open-domain                          \\ \hline
ViHSD \cite{luu2021large}                           & 2021                               & Hate Speech Detection                          & 33,400                             & Open-domain                          \\ \hline
UIT-ViSFD \cite{phan2021sa2sl}                           & 2021                               & Aspect-based Sentiment Analysis                & 11,122                             & Smartphone                           \\ \hline
UIT-ViOCD (Ours)                    & 2021                               & Complaint Detection                            & 5,485                              & Open-domain                          \\ \hline
\end{tabular}
\end{table}
\section{Experiments}
We implement various methods on the UIT-ViOCD dataset and then evaluate its quality by four evaluation metrics such as macro-averaged f1 score, macro-averaged precision, macro-averaged recall, and accuracy \cite{scikit-learn}. We use macro for every metric because our dataset is not imbalanced.
\subsection{Data Pre-processing and Preparation}
We conduct the pre-processing phase to normalize the raw data: lowercasing all texts, removing the HTML tags, extracting white spaces, and applying Unicode standardization. There are many freestyle words in the reviews, like teen code, extending the last sound or acronyms, error typing, or Vietnamese sentences mixed with a few English words. For that reason, we normalize them by accurate words (e.g., "rồii": "rồi" (already)). The old type and the new type are transferred to a unified form (e.g., "òa": "oà"). Moreover, we convert emotional icons to equivalent meaning in Vietnamese. After that, we replace abbreviations with complete words and correct the misspelling. The table \ref{tab:abbreviations} shows some examples. Unlike English, white space between two Vietnamese words is the sign for separating syllables, not words. We tokenize every review using Pyvi library for Vietnamese.

After pre-processing data, we split the dataset into training, validation, and test sets with an 80:10:10 ratio by train\_test\_split function of scikit-learn \cite{scikit-learn}. The results shown in this paper are reported on the test set.

\begin{table}[H]
\centering
\caption{Vietnamese abbreviations in the UIT-ViOCD dataset.}
\label{tab:abbreviations}
\begin{tabular}{ccc}
\hline
\textbf{Abbreviations} & \textbf{Normalization} & \textbf{English meaning} \\ \hline
"dx", "dc", "dk"       & "được"                 & "ok"                     \\
"dt"                   & "điện thoại"           & "phone"                  \\
"god"                  & "tốt"                  & "good"                   \\
"ship"                 & "giao hàng"            & "ship"                   \\ \hline
\end{tabular}
\end{table}
\subsection{Experimental Methods}
\subsubsection{Experimental Methods}We conduct various experiments on different models, including traditional machine learning, deep learning, and transfer learning.

After we convert every review to one hot vector, we encode them mapped into embeddings. In particular, each word is represented by a 64-dimensional word embedding. The embedded representation feeds into three models: Logistic Regression, MLP, and BiLSTM.
\paragraph{Traditional Machine Learning Systems}
      \begin{itemize}
          \item  \textbf{Logistic Regression:} Logistic Regression \cite{hilbe2009logistic} is a binary classification method, as a one-layer neural network with a sigmoid activation function for the prediction task.
       \end{itemize}
\paragraph{Deep Learning Systems}
        \begin{itemize}
             \item  \textbf{MLP (Multi-layer Perceptron):}
 MLP \cite{hornik1989multilayer} has two dense hidden layers (D=512, D=64) with ReLU activation function and the last dense hidden layer (D=1) with sigmoid activation function.               
            \item  \textbf{BiLSTM (Bidirectional Long Short-term Memory):} LSTM was developed and achieved superior performance (Hochreiter and Schmidhuber, 1997) \cite{hochreiter1997lstm}. LSTM is a variation of a recurrent neural network, which has an input gate, an output gate, a forget gate, and a cell. In 2005, Graves et al. \cite{graves2005bidirectional,graves2005framewise} studied BiLSTM in various classification tasks and received good results compared to other neural network architectures. Besides, BiLSTM which consists of two LSTMs has two inputs from forward direction and backward direction. Therefore, BiLSTM can learn contextual information extracted from two directions. Moreover, BiLSTM can learn more features and perform slightly better than LSTM in emotion classification task \cite{zhou2018nlp}. In this paper, we use BiLSTM having 128 units for each.
          
        \end{itemize}
        In addition, we use two pre-training word embedding models (PhoW2V and fastText):
        \begin{itemize}
          \item \textbf{PhoW2V:} PhoW2V is published by Nguyen et al. \cite{nguyen2020phobert}, that were pre-trained on a dataset of 20GB Vietnamese texts. In this paper, we use PhoW2V syllables 300 dims and PhoW2V words 300 dims.
          \item \textbf{fastText:} fastText is released by Grave et al. \cite{grave2018learning}. This model is known for achieving good performance in word performance and sentence classification because they use character-level representations.
           \end{itemize}
\paragraph{Transfer Learning Systems}
        \begin{itemize}
          \item  \textbf{PhoBERT:} Pre-trained PhoBERT models are the first public large-scale monolingual language models pre-trained for Vietnamese (Nguyen et al. 2020) \cite{nguyen2020phobert}. This model approach is based on the same RoBERTa \cite{liu2019roberta}, which eliminates the Next Sentence Prediction (NSP) task of the pre-trained model BERT. There are two PhoBERT versions: base and large. In this paper, we use PhoBERT base and employ RDRSegmenter of VnCoreNLP \cite{vu2018vncorenlp} to pre-process the pre-training data and Transformers\footnote{https://huggingface.co/transformers/} (Hugging Face) for loading pre-trained BERT.
        \end{itemize}
We train the networks using the Adam optimizer by minimizing the binary cross-entropy.
\subsubsection{Results and Discussion} 
As the results in Table \ref{result_model}, most models achieve the art performance state with upper 85\% of the F1-score. The best performing model is PhoBERT with 92.16\% F1-score. These results prove that the dataset has featured enough to help the model classify well.
\begin{table}[H]
\centering
\caption{Results of implemented models on the UIT-ViOCD dataset (\%).}
\label{result_model}
\begin{tabular}{llrrr}
\cline{2-5}
 & \multicolumn{1}{c}{\textbf{Model}} & \textbf{F1-score} & \textbf{Recall} & \textbf{Precision} \\ \cline{2-5} 
 & Logistic Regression                & 89.62             & 88.17           & 91.11              \\ \cline{2-5} 
 & MLP                                & 90.97             & 90.32           & 91.64              \\
 & BiLSTM                             & 89.55             & 92.11           & 87.12              \\
 & BiLSTM + PhoW2V\_words             & 88.46             & 90.68           & 86.35              \\
 & BiLSTM + PhoW2V\_syllables         & 86.36             & 89.61           & 83.33              \\
 & BiLSTM + fasttext                  & 86.33             & 92.83           & 80.69              \\ \cline{2-5} 
 & \textbf{PhoBERT}                   & \textbf{92.16}    & \textbf{90.63}  & \textbf{94.20}     \\ \cline{2-5} 
\end{tabular}
\end{table}
\subsection{Ablation Tests}
We use the ablation technique eliminating structure ingredients for a Vietnamese sentence to evaluate their role in the complaint classification task.
\subsubsection{Punctuation}With the desire to exploit the importance of punctuation marks in Vietnamese complaints, we proceed to include in the dataset model with punctuation and without punctuation to assess. The results are shown in Table \ref{punc_result}, we notice that removing the punctuation marks has affected the performance of the model, significantly reduced results. This experiment proves that punctuation marks contribute to a better classification model.
\begin{table}[!http]
\centering
\caption{The experiment results about punctuation.}
\label{punc_result}
\begin{tabular}{lrr}
\hline
\multicolumn{1}{c}{\multirow{2}{*}{\textbf{Model}}} & \multicolumn{2}{c}{\textbf{F1-score}}                                                     \\ \cline{2-3} 
\multicolumn{1}{c}{}                                & \multicolumn{1}{c}{\textbf{Punctation}} & \multicolumn{1}{c}{\textbf{Without punctation}} \\ \hline
Logistic Regression                                 & 89.62                                   & 89.17                                           \\
MLP                                                 & 90.97                                   & 89.55                                           \\
BiLSTM                                              & 89.55                                   & 88.57                                           \\
BiLSTM + PhoW2V\_syllables                          & 86.36                                   & 86.36                                           \\
BiLSTM + PhoW2V\_words                              & 88.46                                   & 88.06                                           \\
BiLSTM + fastText                                   & 86.33                                   & 85.13                                           \\
\textbf{PhoBERT}                                    & \textbf{92.16}                          & \textbf{91.47}                                  \\ \hline
\end{tabular}
\end{table}

Punctuation is the grammar medium used in writing. It shows the sentence's intonation, expressing nuances about the sentence's meaning and the writer's feelings and attitudes. In specific, question marks differ significantly between complaints and non-complaints comments and mostly appear in complaining reviews. Customers often express their inquiries, the emphasis on the content by using question marks. They desire businesses to explain why the actual product has not been as expected. For example, the review “hàng chính hãng mà bốc hơi quà km là khăn giấy tẩy trang mà không có một lời giải thích hay báo trước là sao?” (“The official item for which the promotional gift is lost is the makeup remover without an explanation, why?”). Besides, users also often use two dots, bullet marks in their complaints. They list the dissatisfaction, express comments with the desire for businesses to improve products and services. Moreover, punctuation is combined to form emotional icons. For example, the most common icons in complaining reviews are ": (" or ": ((" expressing a sad mood, their frustration. Therefore, the models with dataset retaining punctuation have better performance.
\subsubsection{Part-of-speech (POS)}
In this experiment, we use Logistic Regression algorithm with input POS features from the UIT-ViOCD dataset. Then, we conduct repeated experiments eliminating one of the part-of-speech (noun, verb, adjective, and others) from reviews to test the classification performance versus the result of the experiment with input full of POS features. As result shown in Table \ref{pos}, the performance of model for classifying decreased slightly by 1-4\% with F1-score. This result proves that each part-of-speech is a significant feature for predicting complaining reviews.

\begin{table}[H]
\centering
\caption{The result of removing part-of-speech (POS) on the dataset (\%).}
\label{pos}
\begin{tabular}{lrrrr}
\cline{1-3}
 & \multicolumn{1}{c}{\textbf{\begin{tabular}[c]{@{}c@{}}F1-score \end{tabular}}} & \multicolumn{1}{c}{\textbf{\begin{tabular}[c]{@{}c@{}}Accuracy\end{tabular}}} & \multicolumn{1}{c}{\textbf{}} & \multicolumn{1}{c}{\textbf{}} \\ \cline{1-3}
\textbf{Normal} & \textbf{70.09} & 71.83 &  &  \\ \cline{1-3}
\textbf{Removing Verb} & \textbf{67.38} & 69.28 &  &  \\
\textbf{Removing Adjective} & \textbf{66.41} & 68.64 &  &  \\
\textbf{Removing Noun} & \textbf{69.43} & 71.10 &  &  \\
\textbf{Removing Others} & \textbf{69.25} & 71.01 &  &  \\ \cline{1-3}
 & \multicolumn{1}{l}{\textbf{}} &  &  &  \\
 & \multicolumn{1}{l}{\textbf{}} &  &  &  \\
 & \multicolumn{1}{l}{\textbf{}} &  &  &  \\
 & \multicolumn{1}{l}{\textbf{}} &  &  & 
\end{tabular}
\end{table}
\section{Conclusion and Future work}
In this paper, we created the UIT-ViOCD dataset about customer complaints in Vietnamese on open-domain. After building the dataset with 5,485 human-annotated comments on four domains, we implemented various methods for the first evaluations on the dataset.  As a result, we achieved the highest performance by the fine-tuned PhoBERT classifier, a Vietnamese transfer learning model, for identifying complaints with the F1-score of 92.16\%. 

In future, we are going to extend the dataset with new comments on different domains besides the current ones. From that results, we plan to create a tool for identifying complaints automatically. Moreover, the UIT-ViOCD dataset can be extended into many other tasks, such as aspect-based complaint analysis (inspired by the task of aspect-based sentiment analysis \cite{phan2021sa2sl}) or complaint span detection (inspired by the task of toxic span detection \cite{pavlopoulos2021semeval}). Social listening systems can integrate a complaint classification function.

\section*{Acknowledgement}

This research is funded by University of Information Technology-Vietnam National University HoChiMinh City under grant number D1-2021-15.

\bibliographystyle{plain}
\bibliography{Paper}

\end{document}